\documentclass{article}
\usepackage{spconf,amsmath,graphicx}

\usepackage{float}
\usepackage{colortbl}
\usepackage{xcolor}
\usepackage{multirow}
\newcommand{\best}{\cellcolor{red}}
\newcommand{\sbest}{\cellcolor{orange}}
\newcommand{\tbest}{\cellcolor{yellow}}
\definecolor{yellow}{rgb}{1, 1, 0.7}
\definecolor{orange}{rgb}{1, 0.85, 0.7}
\definecolor{red}{rgb}{1, 0.7, 0.7}
\definecolor{normalred}{rgb}{1, 0, 0}
\usepackage{bm}
\usepackage{amssymb}

\newcommand{\bu}{\mathbf{u}}

\newcommand{\nR}{\mathbb{R}}

\newcommand{\cG}{\mathcal{G}}

\newcommand{\cL}{\mathcal{L}}

\title{Sparse2DGS: Sparse-View Surface Reconstruction using 2D Gaussian Splatting with Dense Point Cloud}
\name{Natsuki Takama$^{\dagger}$, Shintaro Ito$^{\dagger}$, Koichi Ito$^{\dagger}$, Hwann-Tzong Chen$^{\dagger\dagger}$, and Takafumi Aoki$^{\dagger}$}
\address{$^{\dagger}$Graduate School of Information Sciences, Tohoku University, Japan\\
$^{\dagger\dagger}$Department of Computer Science, National Tsing Hua University, Taiwan}

\begin{document}
\ninept

\maketitle

\begin{abstract}
Gaussian Splatting (GS) has gained attention as a fast and effective method for novel view synthesis.
It has also been applied to 3D reconstruction using multi-view images and can achieve fast and accurate 3D reconstruction.
However, GS assumes that the input contains a large number of multi-view images, and therefore, the reconstruction accuracy significantly decreases when only a limited number of input images are available.
One of the main reasons is the insufficient number of 3D points in the sparse point cloud obtained through Structure from Motion (SfM), which results in a poor initialization for optimizing the Gaussian primitives.
We propose a new 3D reconstruction method, called Sparse2DGS, to enhance 2DGS in reconstructing objects using only three images.
Sparse2DGS employs DUSt3R, a fundamental model for stereo images, along with COLMAP MVS to generate highly accurate and dense 3D point clouds, which are then used to initialize 2D Gaussians.
Through experiments on the DTU dataset, we show that Sparse2DGS can accurately reconstruct the 3D shapes of objects using just three images.
\end{abstract}

\begin{keywords}
  computer vision, deep learning, 3D reconstruction, multi-view stereo, gaussian splatting
\end{keywords}

\section{Introduction}
\label{sec:intro}

3D reconstruction using multiple images captured from different viewpoints, such as Multi-View Stereo (MVS), is a fundamental technique in computer vision \cite{CV}, and has been applied to digital archiving, VR/AR, etc.
There have been MVS-based methods that use image correspondence through handcrafted features \cite{schoenberger2016mvs} as well as those leveraging deep learning techniques \cite{yao2018mvsnet}.
Deep learning-based approaches achieve higher accuracy in 3D reconstruction than non-learning approaches, while they require a large number of images and their depth maps as ground truth in training.
Both approaches rely on image correspondence across a large number of images; thus, when only a small number of images are used, they can reconstruct only part of the scene, which results in significantly degraded reconstruction accuracy.

With the advent of Neural Radiance Fields (NeRF) for Novel View Synthesis \cite{mildenhall2020nerf}, a 3D reconstruction method called NeuS \cite{wang2021neus} has been proposed to use radiance fields based on Signed Distance Function (SDF) and volume rendering.
Unlike the learning-based MVS method \cite{yao2018mvsnet}, NeuS does not require pre-training or ground-truth depth maps, since radiance fields are optimized using only images for each scene.
On the other hand, NeuS needs a large number of images to optimize the radiance fields for dense and accurate 3D reconstruction.
Several methods have been proposed to fine-tune a pre-trained model using a small number of input images \cite{long2022sparseneus,liang2023retr,na2024uforecon}.
The reconstruction accuracy of these methods is higher than that of NeuS with a small number of input images, while they require a large number of images and several days of training time for the pre-training process.

3D Gaussian Splatting (3DGS) \cite{kerbl3Dgaussians} has been proposed for NVS that represents the radiance fields by a set of ellipsoids (3D Gaussians) with parameters such as color and opacity.
3DGS can optimize the radiance fields faster than NeRF by rendering Gaussians; however, 3DGS cannot be directly used for 3D reconstruction.
Several methods have been proposed to apply 3DGS to 3D reconstruction by adding constraints that allow the Gaussian to represent the 3D shape of an object \cite{Huang2DGS2024,guedon2023sugar}.
While NeRF-based methods require about half a day to optimize the radiance fields, 3DGS-based methods can optimize the radiance fields in about 20 minutes, and the reconstruction accuracy of 3DGS-based methods is the same or higher than that of NeRF-based methods \cite{Huang2DGS2024}.
3DGS-based methods require a large number of images to optimize the radiance fields. The optimization process may fail if there are not enough initial 3D points to serve as Gaussian centers.
Typically, the initial 3D point cloud is obtained by Structure from Motion (SfM), which means that a small number of input images might not provide a sufficient number of 3D points.

As described above, 3D reconstruction from a few viewpoints is a challenging task in MVS.
In addition, since it is difficult to acquire a large number of images in real-world scenarios, it is necessary to perform dense and highly accurate 3D reconstruction of objects from a few viewpoints.
In this paper, we propose a method to reconstruct the precise 3D shape of an object from three input images.
The proposed method uses DUSt3R \cite{Wang_2024_CVPR}, which is a fundamental model for stereo images, to obtain 3D point clouds for each viewpoint.
After removing outliers, the 3D point cloud is integrated with the 3D point cloud obtained by COMAP MVS \cite{schoenberger2016mvs} to create the initial 3D point cloud required in GS.
The proposed method performs 3D reconstruction based on 2DGS \cite{Huang2DGS2024}, which represents the radiance fields by a set of 2D Gaussians.
Through experiments using the DTU dataset \cite{jensen2014large}, we demonstrate the effectiveness of the proposed method in 3D reconstruction from a few viewpoints.

\begin{figure*}[t]
  \centering
  \includegraphics[width=\linewidth]{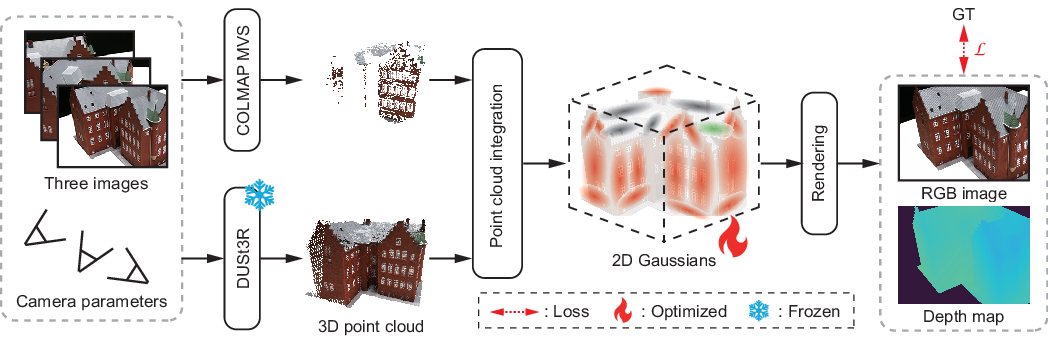}
  \caption{An overview of proposed method.}
  \label{fig:method-overview}
\end{figure*}

\section{Related Work}
\label{sec:work}

This section gives an overview of MVS, neural surface reconstruction, and Gaussian splatting related to this study.

\noindent
{\bf Multi-View Stereo} ---
MVS reconstructs a dense 3D point cloud by estimating depth maps for each viewpoint using correspondence between images and integrating them.
The object surface is reconstructed from the 3D point cloud by using Screened Poisson Surface Reconstruction (SPSR) \cite{kazhdan2013screened}, etc. 
COLMAP \cite{schoenberger2016sfm,schoenberger2016mvs}, which is a representative method, consists of SfM for camera parameter estimation and sparse 3D reconstruction from multi-view images and MVS for dense 3D reconstruction, which is still used as a standard method.
MVSNet \cite{yao2018mvsnet}, which is a deep learning-based MVS method, forms a cost volume using learning-based features and camera parameters, and then inputs it to 3D CNN to estimate depth maps.
There are a lot of methods that have been proposed to extend MVSNet \cite{gu2019cas,zhang2020visibility,chang2022rc}.
All of them assume that multi-view images are input.
When the number of input images is small, the correspondence between images becomes difficult, resulting in a significant decrease in the accuracy of 3D reconstruction.

\noindent
{\bf Neural Surface Reconstruction} ---
The radiance fields and volume rendering of NeRF are used for multi-view 3D reconstruction.
NeuS \cite{wang2021neus} minimizes the error between the rendered image and the ground-truth image, optimizes the SDF-based radiance fields, and reconstructs the object surface from the SDF values.
Unlike the MVS method, which requires ground-truth depth maps during training, NeuS can reconstruct object surfaces only from RGB images.
On the other hand, it requires a long training time for each scene to be reconstructed, resulting in low generalization performance.
In addition, if the number of input images is small, the radiance fields cannot be optimized, and the reconstruction accuracy is degraded.
SparseNeuS \cite{long2022sparseneus} reconstructs object surfaces with high accuracy by fine-tuning trained models for scenes with only a few viewpoint images.
ReTR \cite{liang2023retr} and UFORecon \cite{na2024uforecon} achieve more accurate reconstruction than SparseNeuS \cite{long2022sparseneus} by rendering using Transformer as well as features extracted from images.
However, ReTR \cite{liang2023retr} and UFORecon \cite{na2024uforecon} require a large number of multi-view images and ground-truth depth maps corresponding to each viewpoint for pre-training, and take several days of pre-training.

\noindent
{\bf Gaussian Splatting} ---
3DGS \cite{kerbl3Dgaussians} initializes 3D Gaussians using a sparse 3D point cloud reconstructed by SfM, iteratively optimizes the radiance fields represented by a set of 3D Gaussians, and renders a novel view image using the optimized 3D Gaussians.
Although 3DGS is a fast and accurate method in NVS, 3DGS cannot be used for 3D reconstruction since the radiance fields do not necessarily represent the object's shape.
Focusing on the performance of 3DGS in NVS, some 3D reconstruction methods using 3DGS have been proposed.
GS2Mesh \cite{wolf2024gsmesh} estimates 3D depth maps using stereo matching between stereo images rendered by 3DGS-based NVS.
2DGS \cite{Huang2DGS2024} performs 3D reconstruction by optimizing the radiance fields so that the 2D Gaussians are distributed on the object surface.
These methods achieve fast object surface reconstruction with accuracy comparable to or better than MVS and neural surface reconstruction.
On the other hand, these methods require multiple viewpoint images as input like other methods.
A few viewpoint images not only make it difficult to optimize the radiance fields but also significantly reduce the accuracy of 3D reconstruction since a sufficient number of 3D points for initialization of the 3DGS are not available by SfM.

\section{Method}
\label{sec:method}

As discussed above, to perform 3D reconstruction from a small number of images by taking advantage of the speed and accuracy of GS, it is necessary to obtain dense and highly accurate 3D points from a small number of viewpoints for initialization of Gaussians.
We propose a 3D reconstruction method only from three input images, which employs 2DGS \cite{Huang2DGS2024} with DUSt3R \cite{Wang_2024_CVPR} and COLMAP MVS \cite{schoenberger2016mvs}.
Fig. \ref{fig:method-overview} shows an overview of the proposed method.
The proposed method takes three images and their camera parameters as input.
First, the 3D point clouds are reconstructed from the inputs using COLMAP and DUSt3R, respectively.
Next, the 3D point clouds are integrated after removing outliers to obtain a dense and accurate 3D point cloud.
Then, the radiance fields are optimized using the 3D point cloud as the initial values of 2D Gaussians.
Finally, the depth map for each viewpoint is rendered using the optimized radiance fields and the object surface is reconstructed by integrating them.

\subsection{DUSt3R and COLMAP}

DUSt3R \cite{Wang_2024_CVPR} is used to obtain a dense 3D point cloud from the three images required to initialize Gaussians in 2DGS.
Given an input image $I$ with $W \times H$ pixels, DUSt3R outputs a point map $X \in \nR^{W \times H \times 3}$ that represents the 3D coordinates of each pixel.
Each pixel $X_{l,m}$ in $X$ contains the coordinates of the 3D point corresponding to each pixel $I_{l,m}$ in $I$.
Therefore, a dense 3D point cloud for each viewpoint can be obtained from $X$.
When two or more images are input, multiple stereo image pairs are constructed, and the 3D point cloud is reconstructed from each stereo pair.
Since the coordinate systems of the 3D point clouds are different, they are integrated into a single coordinate system by global alignment of the point maps.
Although a dense 3D point cloud can be obtained from a small number of viewpoint images, DUSt3R cannot provide a 3D point cloud with sufficient accuracy because of the large number of errors in the coordinates of the 3D point clouds.
Therefore, in addition to DUSt3R, the proposed method uses 3D point clouds reconstructed by COLMAP MVS \cite{schoenberger2016mvs}.

\subsection{Point Cloud Integration}

The 3D point clouds obtained by DUSt3R and COLMAP MVS are integrated to have the 3D point cloud for initializing Gaussians in 2DGS.
First, the 3D point cloud from DUSt3R is downsampled with voxels of a specific size.
In this paper, we set this size to 0.005, which was obtained empirically.
Next, 3D points with large distances from neighboring 3D points are considered as outliers and removed using the statistical outlier removal algorithm \cite{Zhou2018}.
Then, the Iterative Closest Point (ICP) algorithm \cite{icp} is used to align the 3D point cloud of DUSt3R with that of COLMAP MVS, resulting in the integration of the two 3D point clouds.
By using the integrated 3D point cloud as the initial value of Gaussians, the optimization of the radiance fields prevents local solutions even when a small number of images are input.

\subsection{2DGS}

2D Gaussians are initialized using the integrated 3D point cloud, and then 2DGS is used to perform  3D reconstruction.
In 2DGS, 2D Gaussians are defined by parameters such as position, color, opacity, etc.
Each coordinate of the integrated 3D point cloud is used as the initial position of 2D Gaussians, and other parameters are initialized in the same way as in 2DGS \cite{Huang2DGS2024}.

3DGS represents the radiance fields as a set of 3D ellipsoids (3D Gaussians), while 2DGS represents the radiance fields as a set of 2D ellipses (2D Gaussians).
A 2D Gaussian used in 2DGS is defined on the local tangent plane, i.e., the $u$-$v$ plane, as follows:
\begin{equation}
  \cG(\bu) = \exp\left(-\frac{u^2+v^2}{2}\right),
\end{equation}
where $\bu=(u,v)$.
In a 3D space defined in the world coordinate system, the 2D Gaussian is defined by
\begin{equation}
  \bm{P}(\bu) = \bm{p} + s_u \bm{t}_u u + s_v \bm{t}_v v,
\end{equation}
where $\bm{p}$ is the center of the 2D Gaussian in a 3D space, $\bm{t}_u$ and $\bm{t}_v$ are tangential vector, and $s_u$ and $s_v$ are scaling factor.
In addition to $\bm{p}$, $\bm{t}_u$, $\bm{t}_v$, $s_u$, and $s_v$, opacity $\alpha$ and color $\bm{c}$ are the parameters of the 2D Gaussian to be optimized.

The color and depth rendering using the radiance fields in 2DGS for a given ray $\bm{r}$ is described in the following.
2D Gaussians on $\bm{r}$ are sorted according to their distance from the camera.
The color $\hat{\bm{C}}$ and depth $\hat{D}$ corresponding to $\bm{r}$ are rendered using alpha blending by
\begin{eqnarray}
  \hat{\bm{C}} &=& \sum_{i=1}^{K} \bm{c}_i\ \omega_i,\\
  \hat{D} &=& \frac{\sum_{i=1}^{K} \omega_iz_i}{\sum_{i=1}^{K}\omega_i + \epsilon},
\end{eqnarray}
respectively, where $i$ is the index number of 2D Gaussians on $\bm{r}$, $K$ is the number of 2D Gaussians on $\bm{r}$, $\epsilon$ is a small number preventing zero division, $\bm{c}_i$, $\alpha_i$, and $z_i$ are color, opacity, and depth of $i$-th Gaussian, respectively.
$\omega_i$ is the weight for alpha blending, which is given by
\begin{equation}
  \omega_i=\alpha_i\cG_i(\bm{u}_i) \prod_{j=1}^{i-1} \left\{1 - \alpha_j\,\cG_j(\bm{u}_i)\right\},
\end{equation}
where $\bm{u}_i$ is the coordinate of the $u$-$v$ plane of the intersection of $\bm{r}$ and the position of the $i$-th 2D Gaussian.

The total loss function used to optimize the radiance fields consists of the color reconstruction loss $\cL_c$ \cite{kerbl3Dgaussians}, the depth distortion loss $\cL_d$ \cite{Huang2DGS2024}, and normal consistency loss $\cL_n$ \cite{Huang2DGS2024}, which is defined by
\begin{equation}
  \cL = \cL_c + \alpha \cL_d + \beta \cL_n,
\end{equation}
where $\alpha$ and $\beta$ are hyperparameters.
We set $\alpha=1000$ and $\beta=0.05$ in this paper.
Following 3DGS \cite{kerbl3Dgaussians}, $\cL_c$ consists of $\cL_1$, which is the L1 loss between the rendered RGB image and the ground truth, and the D-SSIM term, and is given by
\begin{equation}
  \cL_c = (1 - \lambda) \mathcal{L}_1 + \lambda \mathcal{L_{\textrm{D-SSIM}}}.
\end{equation}
The depth distortion loss $\cL_d$ and normal consistency loss $\cL_n$ are used to improve the accuracy of 3D reconstruction as in 2DGS \cite{Huang2DGS2024}, which are given by
\begin{eqnarray}
  \cL_d &=& \sum_{i=1}^{K} \sum_{j=1}^{i-1} \omega_i\omega_j|z_i-z_j|,\\
  \cL_n &=& \sum_{i=1}^{K} \omega_i (1-\bm n_i^\mathrm{T}\bm N),
\end{eqnarray}
where $i$ and $j$ are the index number of 2D Gaussians, $z_i$ and $z_j$ are depth, $K$ is the number of 2D Gaussians, $\bm{n}_i$ is the normal of the $i$-th 2D Gaussian, and $\bm{N}$ is the normal obtained from depth.
Refer to \cite{Huang2DGS2024} for more details.

After optimizing the radiance fields, we render the depth map for each viewpoint and integrate them to reconstruct the mesh model using Truncated Signed Distance Function (TSDF) integration \cite{hernandez2007probabilistic}.

\begin{table*}[t]
    \centering
    \caption{Quantitative results of Chamfer Distance (CD) $\downarrow$ on DTU dataset. Best results are highlighted as \colorbox{red}{1st}, \colorbox{orange}{2nd} and \colorbox{yellow}{3rd}.}
    \label{tab:quantitative-result}
    \resizebox{.98\textwidth}{!}{
    \begin{tabular}{ccccccccccccccccc}
        \hline
        Scan ID & 24 & 37 & 40 & 55 & 63 & 65 & 83 & 105 & 106 & 114 & 118 & 122 & Mean \\
        \hline \hline
        SparseNeuS \cite{long2022sparseneus} & 4.28 & 4.76 & 3.69 & 1.78 & 2.93 & 2.69 & 1.99 & 1.89 & \tbest 1.93  & \tbest 1.12 & 2.24 & 1.88 & 2.60 \\
        ReTR \cite{liang2023retr} & 3.37 & 3.55 & 3.43 & 2.90 & 2.87 & 3.05 & 2.33 & 2.04 & 2.79  & 1.52 & 2.34 & 2.06 & 2.69 \\
        UFORecon \cite{na2024uforecon} & \tbest 1.51 & \best 2.61 & 1.93 & \sbest  1.47 & \best 1.58 & \sbest 1.80 & \best 1.54 & \sbest 1.34 & \best 1.20  & \sbest 0.65 & \best 1.26 & \best 1.25 & \best 1.51 \\
        \hline
        COLMAP \cite{schoenberger2016mvs} & 2.29 & 3.29 & \sbest 1.64 & 2.03 & \tbest 2.33 & 4.51 & 4.48 & 4.02 & 2.47 & 1.87 & 2.52 & 1.80 & 2.77\\
        2DGS \cite{Huang2DGS2024} & ---  & ---  & 5.31 & \tbest 1.60 & 3.57 & 2.29 & --- & 2.73 & 2.20  & 1.36 & 2.20 & \tbest 1.63 & \tbest 2.54 \\
        DUSt3R \cite{Wang_2024_CVPR} & \sbest 1.34 & \tbest 3.28 & \tbest 1.76 & 2.00 & 2.57 & \tbest 2.21 & \sbest 1.85 & \tbest 1.78 & 2.52  & 1.38 & \tbest 1.92 & 2.45 & \sbest 2.09 \\
        Ours & \best 1.01 & \sbest 2.66 & \best 1.56 & \best 1.36 & \sbest 1.83 & \best 1.63 & \tbest 1.91 & \best 1.33 & \sbest 1.53 & \best 0.64 & \sbest 1.29 & \sbest 1.34 & \best 1.51 \\
        \hline
    \end{tabular}
    }
\end{table*}
\begin{table}[t]
  \centering
  \caption{Ablation study of Gaussian initialization methods.
  DUSt3R (Ours) is the 3D point cloud of DUSt3R with the post-processing of the proposed method.}
  \label{tab:quantitative-ablation}
  \begin{tabular}{ccccc}
    \hline
    SfM & MVS & DUSt3R & DUSt3R (Ours) & Mean CD $\downarrow$ \\
    \hline
    $\checkmark$ & $\times$ & $\times$ & $\times$ & 2.54 \\
    $\times$ & $\checkmark$ & $\times$ & $\times$ & 1.90 \\
    $\times$ & $\times$ & $\checkmark$ & $\times$ & 1.76 \\ 
    $\times$ & $\checkmark$ & $\times$ & $\checkmark$ & {\bf 1.51} \\
    \hline
  \end{tabular}
\end{table}

\section{Experiments}
\label{sec:exp}

In this section, we evaluate the performance of the proposed method through the experiments on 3D reconstruction from three input images.

\subsection{Datasets and Metrics}

In this experiment, we use the DTU dataset \cite{jensen2014large}, which is a multi-view image dataset for 3D reconstruction.
DTU consists of 15 scenes, each of which contains 49 or 64 images, ground-truth camera parameters, and ground-truth 3D point clouds.
The size of each image is $1,600 \times 1,200$ pixels, whereas in this experiment, the images are resized to $800 \times 600$ pixels and used as input images.
As for the task of 3D reconstruction from a few viewpoints, we select 3 images that have a small common area among images and are difficult to reconstruct, as in \cite{sparf2023}.
In general, objects that are highly reflective to light are difficult to reconstruct the object shape from a few viewpoint images using any methods.
Therefore, in this experiment, scan 69, 97, and 110, which contain objects with strong reflections, are excluded.

The mesh model reconstructed by each method is evaluated based on the distance between the mesh model and the ground-truth 3D point cloud.
In this experiment, Chamfer Distance (CD) is used as the evaluation metric, which is defined by
\begin{equation}
  \text{CD} = \frac{1}{2N_{pt}} \sum_{i=1}^{N_{pt}} \| \bm{p}_i - \hat{\bm{p}}_i \|_2 + \frac{1}{2N_{pt}^*} \sum_{i=1}^{N_{pt}^{*}} \| \bm{p}^*_i - \hat{\bm{p}}^*_i \|_2,
\end{equation}
where $N_{pt}$ is the number of reconstructed 3D points, $N_{pt}^{*}$ is the number of the ground-truth 3D points, $i$ is the index number of 3D points, $\bm{p}_i$ is the 3D coordinate of $i$-th reconstructed 3D point, $\hat{\bm{p}}_i$ is the 3D coordinate of the $i$-th ground-truth 3D point in the nearest neighbor of the reconstructed 3D point, $\bm{p}^*_i$ is the 3D coordinate of the $i$-th reconstructed 3D point in the nearest neighbor of the ground-truth 3D point, and $\hat{\bm{p}}^*_i$ is the 3D coordinate of the $i$-th ground-truth 3D point.

\subsection{Implementation Details}

We do not perform densification of 2D Gaussians since the proposed method initializes 2D Gaussians using a dense 3D point cloud, and we optimize the parameters of 2D Gaussians obtained in the initialization.
Adam \cite{Kingma2014AdamAM} is used as the optimizer and the number of iterations is 10,000.
The camera parameters used in optimization and 3D reconstruction are the ground-truth camera parameters as in other methods.
We reconstruct the object surface without the background by applying a mask image to the depth map when integrating the rendered depth maps after optimizing the radiance fields.
A single RTX 4090 GPU is used in our experimental environment.
The optimization time of the proposed method is about 100 seconds, and the 3D reconstruction time is about 10 seconds.

\subsection{Baselines}

We demonstrate the effectiveness of the proposed method by comparing it with an MVS-based method: COLMAP \cite{schoenberger2016mvs}, an SDF-based method with pre-training: SparseNeuS \cite{long2022sparseneus}, ReTR \cite{liang2023retr}, and UFORecon \cite{na2024uforecon}, a GS-based method: 2DGS \cite{Huang2DGS2024}, a foundation model: DUSt3R \cite{Wang_2024_CVPR}.
The learning-based MVS methods are not included in the experiments since many studies \cite{long2022sparseneus,liang2023retr,na2024uforecon} have reported low accuracy in 3D reconstruction from a small number of viewpoint images.
UFORecon \cite{na2024uforecon} utilizes a random set training strategy in which pre-training is performed on images from various viewpoints so as to handle large changes in the viewpoints.
Since this paper focuses on the task of 3D reconstruction from a small number of images, UFORecon in this experiment does not use this training strategy in pre-training.

\subsection{Results}

\begin{figure*}[t!]
  \centering
  \includegraphics[width=\linewidth]{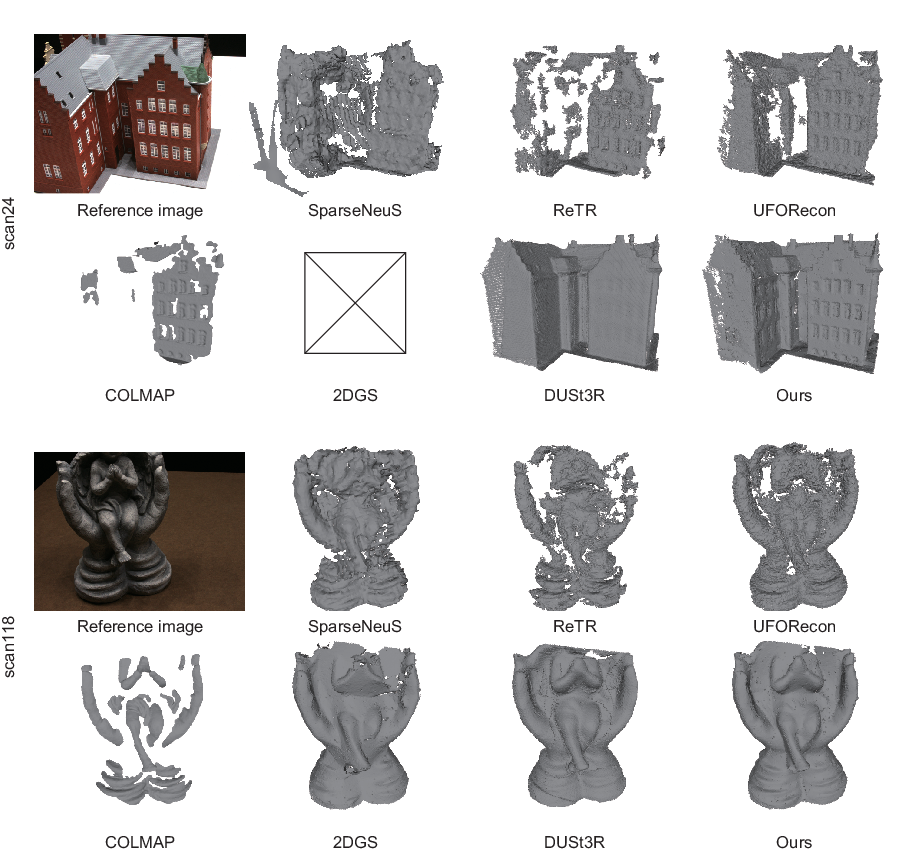}
  \caption{Examples of qualitative results on DTU dataset.}
  \label{fig:qualitative-result}
\end{figure*}

The experimental results are shown in Table \ref{tab:quantitative-result}.
The proposed method exhibits the best or second-best accuracy in almost all scenes.
The number of scenes where the proposed method achieves the best accuracy is comparable to that of UFORecon, which is an existing SOTA method.
As described in Sect. \ref{sec:work}, UFORecon requires a large number of multi-view images and several days of training time for pre-training, whereas the proposed method can reconstruct the object surface in about 100 seconds per scene from only three images.
The proposed method is fast from optimization to reconstruction since the proposed method does not perform densification of Gaussians, but only optimizes the parameters of the dense Gaussians obtained in the initialization.
Fig. \ref{fig:qualitative-result} shows the mesh models reconstructed by each method.
The proposed method can reconstruct the shape of the entire object more accurately than the other methods.
In scan 24, only DUSt3R and the proposed method can reconstruct the shape of the entire object.
DUSt3R cannot reconstruct the detailed shape of the object surface, whereas the proposed method can reconstruct the object surface with high accuracy.
In scan 118, UFORecon is quantitatively the most accurate, however, there are some defects in the reconstruction result.
On the other hand, the proposed method can reconstruct the dense shape of the entire object.

\subsection{Ablation Study}

The main contribution of the proposed method is the initialization of Gaussians using COLMAP MVS \cite{schoenberger2016mvs} and DUSt3R \cite{Wang_2024_CVPR}.
To demonstrate the effectiveness of the proposed Gaussian initialization, we conduct the ablation study on the Gaussian initialization.
COLMAP SfM \cite{schoenberger2016sfm}, COLMAP MVS \cite{schoenberger2016mvs}, and DUSt3R \cite{Wang_2024_CVPR} are compared with the proposed method for acquiring a 3D point cloud for the Gaussian initialization.
Table \ref{tab:quantitative-ablation} shows the experimental results.
COLMAP MVS \cite{schoenberger2016mvs} and DUSt3R  \cite{Wang_2024_CVPR} improve the reconstruction accuracy compared to COLMAP SfM \cite{schoenberger2016sfm}, which is generally used in GS.
The proposed method has the highest accuracy, demonstrating the effectiveness of the proposed method in the Gaussian initialization.

\section{Conclusion}
\label{sec:con}

We have presented a 2DGS-based method, referred to as {\it Sparse2DGS}, that requires only three input images for 3D reconstruction.
In Sparse2DGS, we utilize DUSt3R \cite{Wang_2024_CVPR} and COLMAP MVS \cite{schoenberger2016mvs} to obtain the dense 3D point clouds for initializing 2D Gaussians in 2DGS.
Through experiments using the DTU dataset \cite{jensen2014large}, we have demonstrated that Sparse2DGS can achieve higher reconstruction accuracy compared with conventional methods.
In addition, Sparse2DGS takes about 100 seconds in 3D reconstruction, while the SOTA method such as UFORecon \cite{na2024uforecon} requires a large number of multi-view images and several days of training time for pre-training.

\section{Acknowledgment}
This work was supported in part by JSPS KAKENHI 23H00463 and 25K03131.

{\small
  \bibliographystyle{IEEEbib}
  \bibliography{strings,refs}
}
\end{document}